\documentclass[pdflatex,sn-mathphys-num] {sn-jnl}

\usepackage{graphicx}%
\usepackage{multirow}%
\usepackage{amsmath,amssymb,amsfonts}%
\usepackage{amsthm}%
\usepackage{mathrsfs}%
\usepackage[title]{appendix}%
\usepackage{xcolor}%
\usepackage{textcomp}%
\usepackage{manyfoot}%
\usepackage{booktabs}%
\usepackage{algorithm}%
\usepackage{multirow}
\usepackage{natbib}
\usepackage{algpseudocode}%
\usepackage{listings}%
\usepackage{dcolumn}
\usepackage{bm}

\theoremstyle{thmstyleone}%
\theoremstyle{thmstyletwo}%

\theoremstyle{thmstylethree}%

\DeclareUnicodeCharacter{2217}{\textasteriskcentered}

\begin{document}


\title[Article Title]{PERM EQ x GRAPH EQ: Equivariant Neural Networks for Quantum Molecular Learning}





\author[1]{Saumya Biswas}\email{sbiswas4@umd.edu}

\author[2]{Jiten Oswal}\email{jiten.p.oswal@gmail.com}

\affil[1]{Independent Researcher, College Park, Maryland, United States}

\affil[2]{Independent AI Researcher, San Francisco, California, United States}



\abstract{In hierarchal order of molecular geometry, we compare the performances of Geometric Quantum Machine Learning models. Two molecular datasets are considered: the simplistic linear shaped $LiH$-molecule and the trigonal pyramidal molecule $NH_3$. Both accuracy and generalizability metrics are considered. A classical equivariant model is used as a baseline for the performance comparison. The comparative performance of Quantum Machine Learning models with no symmetry equivariance, rotational and permutational equivariance, and graph embedded permutational equivariance is investigated. The performance differentials and the molecular geometry in question reveals the criteria for choice of models for generalizability. Graph embedding of features is shown to be an effective pathway to greater trainability for geometric datasets. Permutational symmetric embedding is found to be the most generalizable quantum Machine Learning model for geometric learning.}




\maketitle

\section{Introduction}\label{sec1}
In the Noisy Intermediate Scale Quantum (NISQ) era \cite{preskill2018quantum}, Quantum Machine Learning (QML) has remained a theoretical prospect because of the high computational cost of the measurement cost and optimization iterations \cite{liu2023can,chinzei2025trade,schuld2022quantum}. Parametrized Quantum Circuits (PQC) is the heart and soul of Variational Quantum Algorithms (VQA) \cite{cerezo2021variational} that iteratively optimizes the parameters for minimization of an objective function. A particular approach called Geometric Quantum Machine Learning (GQML) exploit the knowledge of symmetry of data to improve the machine learning task \cite{chinzei2024resource}. Symmetry Equivariant Quantum Neural Networks (EQNN) have emerged as the workhorse that performs the best in terms of generalization and trainability \cite{bronstein2021geometric,verdon2019quantum,zheng2023speeding,larocca2022group,meyer2023exploiting,ragone2022representation,nguyen2024theory,sauvage2024building}. In 3D physics and chemistry problems, recognizing the symmetric properties through the neural network design has become an important trick of the trade \cite{shen2024interpretable,feng2023learning,smidt2021euclidean} EQNN may be a promising pathway with superior generalization and convergence since it utilizes the spatial geometrical information and symmetry operations. These EQNN models consider the node features (non-geometric information) \cite{yan2023quantum}, node coordinates (geometric information) \cite{baek20223d}, or them simultaneously \cite{liu2025rotation}. The approach coincides with a 3D graph based approach as spatial coordinates readily possess the structure of graphs. The goals and challenges are in fact shared between Quantum Graph Neural Networks (QGNN) and their classical counterparts in learning three dimensional graph data 
\cite{skolik2023equivariant,chinzei2024resource}.

In VQA, an optimized parametrization of a model is sought that entails an inductive bias relevant to the learning task. A symmetrization process on the gateset representing the network can produce an equivariant gateset \cite{meyer2023exploiting}. Such equivariant quantum circuits have emerged as a tool for highly trainable and generalizable model for QML. While most strategies exploit rotational symmetries and rotational and permutational symmetries at most, graph embedding the features can lead to further unravelling of the symmetries of the problem. This is the central question of this work. If tapping into graph permutational equivariance can improve the generalizability or performance of a GQML problem over most commonly used rotational and permutational equivariance. While a simple minded guess for the answer would be affirmative, since tapping into more symmetries should help generalizability, we want to investigate the question rigorously with specific examples. Additionally, we want to understand the scenarios or problems where such advantages would be more prominent.

\section{Properties of the dataset}
In this work, we investigate the generalizability of QML networks on ab initio quantum chemistry data. There are classical codes such as PSI4 \cite{turney2012psi4}, ORCA \cite{neese2020orca}, Gaussian \cite{frisch2009gaussian} etc. that produces such datasets. We have used PSI4 to create our dataset. PSI4 relies upon Density Functional Theory (DFT) for creation of dataset. Commonly, it solves for the groundstate electronic density using Kohn-Sham equations \cite{bickelhaupt2000kohn}. PSI4 builds upon the Hartree-Fock approximation usually done with a single Slater determinant. It improves the calculations with second-order perturbation theory and Coupled Cluster Single and Double (CCSD) framework. The dataset includes positions, energy, and force. PSI4 utilizes analytical gradients and Hessians for the force calculation from the quantum mechanical electronic energy. It does a proper accounting of the quantum potential energy surface. The energy calculation comes from a formula akin to $E(\mathbf{R})= \langle  \Psi(\mathbf{R}) \hat{H}_{electronic} \Psi(\mathbf{R}) \rangle$, calculated from groundstate calculation ($\mathbf{R}$ being the position vector and $\hat{H}$-being the Hamiltonian operator). The force is calculated from a formula $\mathbf{F}_i = -\frac{\partial E(\mathbf{R})}{\partial \mathbf{R}_i}$. For simplicity, we only use the Restricted Hartree-Fock method, with geometry scans and energy/force surfaces for the creation of our data. 

\begin{table}[htbp] 
\centering
\small
\begin{tabular}{|p{2.4cm}|p{3.0cm}|p{3.0cm}|p{3.5cm}|}
\hline
\textbf{Variable} & \textbf{Description} & \textbf{Data Type / Units} & \textbf{Array Shape} \\
\hline
\texttt{Positions} 
& Cartesian coordinates of all atoms in NH$_3$ (N + 3 H) 
& Float64, in \AA{}ngstr\"oms (\AA{}) 
& $(2400,\ 4,\ 3)$  
\\
\hline
\texttt{Forces} 
& Nuclear forces computed from the electronic structure calculation 
& Float64, in eV/\AA{} (or Hartree/Bohr depending on PSI4 settings) 
& $(2400,\ 4,\ 3)$  
\\
\hline
\texttt{Energy} 
& Total electronic energy for each molecular configuration 
& Float64, in Hartree (1 Ha = 27.211386 eV)  
& $(2400)$  
\\
\hline
\texttt{Atom Index 0} 
& Nitrogen (N) atom position and force entries 
& Same units as above 
& Slices \texttt{[:,0,:]}  
\\
\hline
\texttt{Atom Indices 1--3} 
& Hydrogen atoms H$_1$, H$_2$, H$_3$ 
& Same units as above 
& Slices \texttt{[:,1:4,:]} 
\\
\hline
\texttt{Coordinates (x,y,z)} 
& Spatial components of each atom 
& Float64, \AA{}  
& Innermost dimension \texttt{(3)}  
\\
\hline
\texttt{Force Components} $(F_x, F_y, F_z)$
& Components of nuclear forces for each atom 
& Float64, eV/\AA{}  
& Innermost dimension \texttt{(3)}  
\\
\hline
\end{tabular}
\caption{Complete specification of the NH$_3$ dataset generated with PSI4. Each sample contains atomic positions, electronic energy, and nuclear forces obtained from ab initio electronic structure calculations.}
\end{table}

\begin{table}[h!]
\centering
\small
\begin{tabular}{|p{2.6cm}|p{4.0cm}|p{4.3cm}|}
\hline
\textbf{Feature} & \textbf{Mathematical Definition} & \textbf{Description / Use in Model} \\
\hline

Bond vector $\mathbf{r}_{NH_i}$ 
& $\mathbf{r}_{NH_i} = \mathbf{x}_{H_i} - \mathbf{x}_N$ 
& Relative vector from nitrogen to each hydrogen; used for geometry-dependent rotations and message passing. \\
\hline

Bond distance $d_i$
& $d_i = \|\mathbf{r}_{NH_i}\|$ 
& N–H bond length; RY/RZ rotations in the quantum embedding circuit are controlled by the length. \\
\hline

Unit bond direction $\hat{\mathbf{r}}_{NH_i}$
& $\hat{\mathbf{r}}_{NH_i} = \mathbf{r}_{NH_i}/\|\mathbf{r}_{NH_i}\|$ 
& Directional information required for rotation-equivariance; used in angular encoding. \\
\hline

Bond angle $\theta_{ij}$
& $\displaystyle 
\theta_{ij} = \arccos 
\left(
\frac{\mathbf{r}_{NH_i} \cdot \mathbf{r}_{NH_j}}
{\|\mathbf{r}_{NH_i}\|\;\|\mathbf{r}_{NH_j}\|}
\right)
$ 
& Angle between N–H bonds (H$_i$–N–H$_j$); used for cross-bond geometric coupling via RZ rotations. \\
\hline

Pairwise geometric interaction
& $\big(\mathbf{r}_{NH_i},\;\mathbf{r}_{NH_j}\big)$  
& Pairwise structural coupling between bond directions; used for cross-bond entanglement patterns. \\
\hline

Cross-bond entanglement structure
& CNOTs applied between qubits for different bonds (e.g., $1\!\leftrightarrow\!2$, $3\!\leftrightarrow\!4$)
& Quantum circuit encoding multi-body correlations. \\
\hline

Normalized angle features
& $\sin\theta_{ij},\ \cos\theta_{ij}$ (implicitly through RZ($\theta_{ij}$))
& Implicit feature embedding through trigonometric functions of rotation gates. \\
\hline

\end{tabular}
\caption{\label{table_graph_QML}
Derived geometric features computed from PSI4 atomistic positions and used in the Graph Permutaionally Equivariant QML force/energy model for NH$_3$. These features encode distances, directions, angles, and pairwise many-body structure necessary for SE(3)-equivariant learning.
}
\end{table}

Now, in general, the quantum chemistry codes do not produce data that inherit the symmetry of the problem. For example, the water molecule $H_2 O$ has two Hydrogen atoms that should be indistinguishable. Physically, the environments of the two Hydrogen atoms can be different and the force calculated on them may be different too \cite{seki2020bending,kresse1996efficiency},\cite[chapter-11]{martin2020electronic}. The customary datasets obtained from quantum chemistry software will not necessarily have Hydrogen atoms symmetrically placed around the Oxygen atom (expected from $C_{2v}$-symmetry). Almost all samples in a dataset comes with an asymmetric geometry characteristic of thermal-like fluctuations \cite{yang2019deconstructing}.

\section{The orders of symmetry}
Now, we entertain the question of suitability of method for quantum learning the ab initio Hartree-Fock data. We have already discussed the lack of certain symmetries in the data e.g. translational, rotational, and permutational. A natural question is whether the lack of symmetry in the data hinders the use of a symmetry equivariant network \cite{wang2022approximately}. The answer is no, but the model architecture must handle the symmetries correctly. While the data might lack a certain symmetry, the model can learn much better from an assumption of the expected symmetry \cite{wang2022approximately}. It has been found that models may struggle to discover symmetries only from data \cite{perin2025ability}. Therefore giving the model the information explicitly would be the more effective route to generalization. Now we discuss why such a symmetry is desired in a QML model.

QML depends on smooth parametrization of quantum circuits with well behaved gradients. Therefore a model with a Lie Group symmetry is ideal, especially for the purpose of predicting differentiable physics (force being the negative gradient of energy). Lie groups are in some sense a continuous set generalization of groups. The dataset in question include 3D coordinates and geometry, therefore the Lie group of E(3) (the full Euclidean group in 3D \cite{satorras2021n}) may be the proper choice for well functioning quantum circuits. We can feed the data into the circuit in relative distances $r_{ij}=|\mathbf{r}_{i} - \mathbf{r}_{j}|$, effectively making the quantum circuit layers translation equivariant. We discuss the specific rotational equivariant encoding we choose in the next section. The objective of the paper is to investigate the performance improvement that possibly comes from additional symmetry choices in the equivariance. We are especially interested in the performance differential that the final addition-- a permutational equivariance may bring. The physical principle of the molecule embodies permutational symmetry even if the data may lack it. The choice of a Lie group symmetry is motivated by the need for a smooth structure that a continuous, differentiable manifold brings.

\section{SYMMETRIES INDUCED BY DATA EMBEDDINGS}
Symmetries have a fundamental role both in physics \cite{meyer2023exploiting} and geometric deep learning \cite{bronstein2021geometric}. In Variational Quantum Eigensolvers \cite{seki2020symmetry}, the symmetries of the ansatze are derived from the physical geometry \cite{barron2021preserving}. And a machine learning task might entail a prediction $y \in Y$ associated with data points $x \in X$ that is invariant under a symmetry group S with representation $V_s: S \rightarrow  Aut(\chi) $ if $y(Vs[x]) = y(x)$ for all $x \in X$ and $s \in S$. Since, a symmetry invariant model does not distinguish between datapoints related by symmetry transformations, it restricts the model's predictions to meaningful symmetric ones. In the relevant QML, we map the symmetry operations of the data onto the unitary transformations of the qubit's Hilbert space. The natural description of these encodings are the representation theories of the relevant spin Lie group. A representation of a group is a map $U: S \rightarrow  Aut(V), s \rightarrow U_s $ that respects the group composition law $U_{s_1 \otimes s_2}=U_{s_1}U_{ s_2}$. In QML, the relevant $U_{s}$ are unitary and $V$ has an inner product defined making U a unitary representation.

Lie Algebra SO(3) and SU(2) are isomorphic, and the 3D graph embedding of the spatial information of the molecule can be likewise embedded into the generators of the qubit gates. Lie groups are also smooth manifolds with group composition and inversions being smooth maps. Lie algebras are the generators of elements of the Lie group. The associated algebra for a matrix Lie group $\mathcal{S}$ is,
\begin{eqnarray}
\mathfrak{s}=\operatorname{Lie}(\mathcal{S})=\{G: \exp (t G) \in \mathcal{S} \text { for all } t \in \mathbb{R}\}    
\end{eqnarray}
As continuous groups with a differentiable structure, Lie groups are the framework for dealing with continuous symmetries.

\subsection{Equivariance vs Invariance}
Let us assume that the problem can be represented on a graph, $G=(V,E)$ with a adjacency matrix A (V,E being the set of vertices and edges respectively). Since, a simultaneous permutation of rows and columns lead to the same graph, it is reasonable to require that the quantum state encoding of the problem should inherit the property. To this end, we shall have to define a symmetry called label symmetry which represents this permutational symmetry.

But, before that we quickly review the definitions of symmetry invariance and equivariance. We do this with respect to permutational symmetry (which is very important for this work), the definition is generalizable to symmetries other than permutational symmetry. If the quantum state embedding of the adjacency matrix A is invariant under a permutation $\pi$ of the rows or columns of A,
\begin{eqnarray}
    |\phi(A) \rangle = |\phi(\pi(A)) \rangle \ \ \ \ \ \ \label{eq_inv_state}
\end{eqnarray}
then the quantum circuit embedding is termed permutation invariant. The less restricted condition for equivariance is,
\begin{eqnarray}
    |\phi(A) \rangle = P_{\pi}|\phi(\pi(A)) \rangle \ \ \ \ \ \ \label{eq_equi_state}
\end{eqnarray}

We now switch to a density matrix $\rho$-formalism since that is the more suitable quantum dynamical framework for qubits \cite{coyle2025training} and define the label symmetry we shall need. If $\rho$ is the input density matrix and f is the function to be learned, the label symmetry G requires that
\begin{eqnarray}
f(\rho)=f\left(U_g \rho U_g^{\dagger}\right), \quad \quad \forall g \in G, \forall \rho,     \ \ \ \ \ \label{eq_label_symm}
\end{eqnarray}

where $U_g$ is a unitary representation of the group $G$. We can ensure the equivariance of the QNN by imposing the label symmetry on the quantum circuit. A parametrized quantum circuit $U(\boldsymbol{\theta})$, invariant under the action of $G$ is defined to be a Equivariant QNN \cite{chinzei2024resource}:

\begin{eqnarray}
\left[U(\boldsymbol{\theta}), U_g\right]=0 \quad \forall g \in G .
\end{eqnarray}

As a consequence of this symmetry, the input and output density matrix is related by the equivariance condition,
\begin{eqnarray}
    \rho_{in} = U_g \rho_{out} U_g^{\dagger} \ \ \ \label{eq_equivariance_density}\\
    \rho_{out} = U(\boldsymbol{\theta}) \rho_{in} U^{\dagger}(\boldsymbol{\theta}) \ \ \ \ \ \label{eq_equivariance_op}
\end{eqnarray}
Eq. \ref{eq_equivariance_density} is the density matrix generalization of the ket equivariance condition Eq. \eqref{eq_equi_state} and follows from Eq. \eqref{eq_label_symm} directly ($U(\boldsymbol{\theta})\left[U_g \rho U_g^{\dagger}\right] U^{\dagger}(\boldsymbol{\theta})= U_g\left[U(\boldsymbol{\theta}) \rho U^{\dagger}(\boldsymbol{\theta})\right] U_g^{\dagger}$). Additionally, for a $G$-symmetric observable $O$ (i.e., $\left[O, U_g\right]=0$ ), the label symmetry of Eq. \eqref{eq_label_symm} can be found for the function, $f(\rho)=\operatorname{tr}\left[U(\boldsymbol{\theta}) \rho U^{\dagger}(\boldsymbol{\theta}) O\right]$.

We now consider the practical parametrized quantum circuit that fits the bill in VQA. Because of no cloning theorem, the classical data has to be fed into the network multiple times, the central idea of the data reuploading model \cite{meyer2023exploiting,perez2020data}. In our work, the data-embedding unitary $U_d(\boldsymbol{x}), d=\{1,2,...,d\}$ is interleaved with parametrized quantum circuits $U(\boldsymbol{\theta})$ as

\begin{eqnarray}
|\psi(\boldsymbol{\theta}, \boldsymbol{x})\rangle=  U(\boldsymbol{x}) U_D(\boldsymbol{\theta}) \ldots U_1(\boldsymbol{\theta}) U(\boldsymbol{x}) U_0(\boldsymbol{\theta}) U(\boldsymbol{x}) \left|\psi_0\right\rangle,
\end{eqnarray}
where $ |\psi_0 \rangle $ is the initial state prepared according to symmetry equivariance or invariance requirements.

With the data encoding and parametric layers, the condition for equivariance can include the data and parameters explicitly,
\begin{eqnarray}
U(\boldsymbol{\theta}, V_g[\boldsymbol{x}])=U_g U(\boldsymbol{\theta}, V_g[\boldsymbol{x}]) U_g^{\dagger}, \ \ \ \forall \boldsymbol{\theta}, \boldsymbol{x}, g \in G \ \ \ \ \ \label{eq_equiv_ckt}
\end{eqnarray}
where $V_g$ is the representation for $g \in G$, the symmetry group.

\subsection{Rotational Equivariance}
Every Lie group can be represented on a large enough quantum system \cite{meyer2023exploiting}. This is a consequence of the mathematical fact that all compact Lie group is a subgroup of a large unitary group, which in turn, is a subgroup of a larger special unitary group \cite{meyer2023exploiting,morales2020universality}.

We briefly review how an embedding that is equivariant with respect to the special orthogonal group in 3 dimensions, $\mathrm{SO}(3)$, can be constructed. For the methodology of the general task, see \cite{meyer2023exploiting}. Elements of the Lie group $\mathrm{SO}(3)$ are all possible rotations of a sphere centered at the origin. We refer to them with lower case letters to distinguish them from quantum mechanical gates (to be labeled with upper case letters). We denote as $r_x(\alpha), r_y(\alpha)$ and $r_z(\alpha)$ the three canonical rotations about the axes of the coordinate system. This parametrization is oftentimes written in terms of the Euler angles given by
\begin{eqnarray}
r(\psi, \theta, \phi)=r_z(\psi) r_x(\theta) r_z(\phi)  
\end{eqnarray}

In terms of the vector of Pauli operators $\boldsymbol{\sigma}=(X, Y, Z)$, we can have an equivariant embedding of a data point $\boldsymbol{x}=(x_1,x_2,x_3) \in \mathbb{R}^3$ as
\begin{eqnarray}
U(\boldsymbol{x})=e^{-\frac{i}{2}\left(x_1 X+x_2 Y+x_3 Z\right)}=e^{-\frac{i}{2}\langle\boldsymbol{x}, \boldsymbol{\sigma}\rangle} \ \ \ \ \label{eq_One_Qub_rot_E}   
\end{eqnarray}
It is essentially a mapping from data points to the Bloch sphere. Using the fact that conjugation with a Pauli rotation leads to the induced representation of rotations,
\begin{eqnarray}
\begin{aligned}
\left\langle r_x(\alpha) \boldsymbol{x}, \boldsymbol{\sigma}\right\rangle & =\left\langle\boldsymbol{x}, r_x(-\alpha) \boldsymbol{\sigma}\right\rangle \\
& =R_X(-\alpha)\langle\boldsymbol{x}, \boldsymbol{\sigma}\rangle R_X(\alpha), \\
\left\langle r_z(\alpha) \boldsymbol{x}, \boldsymbol{\sigma}\right\rangle & =R_Z(-\alpha)\langle\boldsymbol{x}, \boldsymbol{\sigma}\rangle R_Z(\alpha),
\end{aligned}    
\end{eqnarray}

to deduce that

\begin{eqnarray}
\begin{aligned}
U(r(\psi, \theta, \phi) \boldsymbol{x})= & R_Z(-\psi) R_X(-\theta) R_Z(-\phi) e^{-\frac{i}{2}\langle\boldsymbol{x}, \boldsymbol{\sigma}\rangle} 
 R_Z(\phi) R_X(\theta) R_Z(\psi) \\
= & U(-\psi,-\theta,-\phi) U(\boldsymbol{x}) U^{\dagger}(-\psi,-\theta,-\phi)
\end{aligned} \ \ \ \ \label{eq_equiv_SU2}
\end{eqnarray}
which is Eq. \eqref{eq_equiv_ckt} i.e. an equivalent statement of the equivariance condition (Eq. \eqref{eq_equivariance_op}) in the Heisenberg picture of operators.

Eq. \eqref{eq_equiv_SU2} uses the parametrization for arbitrary single qubit gates (unitaries) $U \in \mathrm{SU}(2)$, in terms of three angles,\begin{eqnarray}
    U(\psi, \theta, \phi):=R_Z(\psi) R_X(\theta) R_Z(\phi)
\end{eqnarray}

We also consider two-qubit rotationally invariant operator and observable,
\begin{eqnarray}
    H^{(i,j)}(J) = -J \left( X^{(i)}X^{(j)} + Y^{(i)}Y^{(j)} + Z^{(i)}Z^{(j)} \right) \ \ \ \ \label{eq_Two_Qub_Op}\\
    O = X^{(i)}X^{(j)} + Y^{(i)}Y^{(j)} + Z^{(i)}Z^{(j)}  \ \ \ \ \label{eq_Two_Qub_Obs}
\end{eqnarray}
Eq.s \eqref{eq_Two_Qub_Op} and \eqref{eq_Two_Qub_Obs} are reminiscent of the Heisenberg Hamiltonian- which is rotationally invariant. For initialization in a rotationally invariant system, the two qubit state ``singlet state'' is a suitable choice.
\begin{eqnarray}
    |\psi_{singlet} \rangle = \frac{1}{\sqrt{2}} \left( | 0 1 \rangle - | 1 0 \rangle \right) \ \ \ \ \ \label{eq_singlet_state}
\end{eqnarray}

\subsection{\label{sec_Perm_Eq} Permutational Equivariance}
We now want to encode a permutational symmetry of active atoms (The(two) Hydrogen atom(s) in LiH($NH_3$)) in the molecules we investigate. Notwithstanding the lack of permutational symmetry of the data, our trained model and predictions would be symmetrical. This symmetry group will be a subgroup of the permutation group (or symmetric group) $S_d$ of the $d$ different data features $\boldsymbol{x}=\left(x_1, \ldots, x_d\right)$.

Generating each permutation (the reshufflings denoted by $\sigma$) over the data feature is an element of the symmetric group of the data features
\begin{eqnarray}
\sigma\left(x_1, x_2, \ldots, x_d\right)=\left(x_{\sigma(1)}, x_{\sigma(2)}, \ldots, x_{\sigma(d)}\right) .    
\end{eqnarray}

The permutation operations can be constructed by concatenating the transpositions ($\tau$, exchanging two data features). Each $\sigma \in S_d$ can be written as $\sigma=\tau_m \ldots \tau_2 \tau_1$ for some transpositions $\left\{\tau_i\right\}$.

Following \cite{skolik2023equivariant}, our choice for the quantum embedding is (based on the graph $G(V, E)$ with node features $\boldsymbol{\alpha}$ and weighted edges $\mathcal{E}$), 

\begin{eqnarray}
\begin{aligned}
|\mathcal{E}, \boldsymbol{a}, \boldsymbol{\beta}, \boldsymbol{\gamma}\rangle_p= & U_N\left(\boldsymbol{a}, \beta_p\right) U_G\left(\mathcal{E}, \gamma_p\right) \\
& \ldots U_N\left(\boldsymbol{a}, \beta_1\right) U_G\left(\mathcal{E}, \gamma_1\right)|s\rangle,
\end{aligned} \ \ \ \ \label{eq_gra_per_enc}
\end{eqnarray}

where $\boldsymbol{\beta}, \boldsymbol{\gamma} \in \mathbb{R}^p$ are trainable parameters, the n-qubit state $|s\rangle$ is the uniform superposition of bit-strings of length $n$,
\begin{eqnarray}
|s\rangle=\frac{1}{\sqrt{2^n}} \sum_{x \in\{0,1\}^n}|x\rangle,    
\end{eqnarray}

The encoding operations are explicitly,
\begin{eqnarray}
U_N\left(\boldsymbol{\alpha}, \beta_j\right)=\bigotimes_{l=1}^n \operatorname{Rx}\left(\alpha_l \cdot \beta_j\right), \ \ \ \ \operatorname{Rx}(\theta)=e^{-i \frac{\hat{2}}{2} X}   \\
U_G\left(\mathcal{E}, \gamma_j\right)=\exp \left(-i \gamma_j H_{\mathcal{G}}\right), \ \ \ \ H_{\mathcal{G}}=\sum_{(i, j) \in \mathcal{E}} \varepsilon_{i j} \sigma_z^{(i)} \sigma_z^{(j)},
\end{eqnarray}
with $\mathcal{E}$ are the edges of graph $\mathcal{G}$ weighted by $\varepsilon_{i j}$. The first part $U_G$ of each alternating layer encodes edge features, and the second part $U_N$ encodes node features. Each part is parametrized by one parameter {$\beta_l, \gamma_l$, respectively}.

Let list of edges $\mathcal{E}$ have the corresponding edge weights $\varepsilon_{i j}$, and node features $\boldsymbol{a} \in \mathbb{R}^n$ with $n=|\mathcal{V}|$. Let $\sigma$ be a permutation of the vertices in $\mathcal{V}, P_\sigma \in \mathbb{B}^{n \times n}$ and the corresponding permutation matrix that acts on the weighted adjacency matrix $A$ of $\mathcal{G}$, and $\widetilde{P}_\sigma \in \mathbb{B}^{2^n \times 2^n}$ be a matrix that maps the tensor product $\left|v_1\right\rangle \otimes\left|v_2\right\rangle \otimes \ldots \otimes\left|v_n\right\rangle$ with $\left|v_i\right\rangle \in \mathbb{C}^2$ to $\left|v_{\tilde{p}_\sigma(1)}\right\rangle \otimes\left|v_{\tilde{p}_\sigma(2)}\right\rangle \otimes \ldots \otimes\left|\tilde{v}_{\tilde{p}_\sigma(n)}\right\rangle$. Then, it was shown in \cite{skolik2023equivariant}

$$
\left|\mathcal{E}_A, \boldsymbol{\alpha}, \boldsymbol{\beta}, \boldsymbol{\gamma}\right\rangle_p=\tilde{P}_\sigma\left|\mathcal{E}_{\left(P_\sigma^T A P_\sigma\right)}, P_\sigma^T \boldsymbol{\alpha}, \boldsymbol{\beta}, \boldsymbol{\gamma}\right\rangle_p,
$$
with $\mathcal{E}_{(\cdot)}$ being a specific permutation of the adjacency matrix $A$. Therefore, the quantum embedding in Eq. \eqref{eq_gra_per_enc} does satisfy the equivariance condition, eq. \eqref{eq_equi_state}.

\section{Methods}
We are going to compare the performance of four methods in performing the same learning tasks. Their names are chosen to represent the most important characteristic of each method although they are composite methods (using tricks from more than one class) to be fair. For example, the ``Rotationally Equivariant method'' below uses permutational equivariance as well, only not as much as the ``Graph Permutation Equivariant QML'' method, which additionally represents the data on a graph. We briefly describe the four methods below:

\begin{itemize}
    \item \textbf{Rotationally Equivariant QML:} By means of SO(3) equivariant rotations (Eq. \eqref{eq_One_Qub_rot_E}), atomic positions are encoded  on singlet-initialized qubit pairs (Eq. \eqref{eq_singlet_state}), aided by Heisenberg-like interactions coupling the qubits (Eq.s \eqref{eq_Two_Qub_Op}, \eqref{eq_Two_Qub_Obs}) across variational layers. Training includes a warm-up when force weight is gradually increased. It stabilizes the learning of the energy surface prior to adding force constraints.
    
    \item \textbf{Non-Equivariant QML:} This is the raw method without symmetry equivariance. $R_Y$/$R_Z$ rotations take in the raw interatomic distances directly. CNOT gates perform the entanglement between qubits. The training is also chosen to be naive. From the start of training, both energy and force losses are optimized.
    
    \item \textbf{Graph Permutation Equivariant QML:} This is the sophisticated method where molecular data is first represented as features of a graph (Table \ref{table_graph_QML} (and \ref{table_NH3} for $NH_3$)). The encoding of the graph features is done permutationally equivariantly (Subsection \ref{sec_Perm_Eq}). Qubit pairs represent bonds, with bond distances and angles (encoded with learnable weights) across these qubit pairs. Cross-bond CNOT gates codify inter-bond correlations. In a two-phase training, energy is optimized first for (~200) epochs, then force gradients are added in the subsequent (~200) epochs.
    
    \item \textbf{Classical Rotationally Equivariant NN:} This is a purely classical method that we compare the QML performance against. A Multi-Layer-Perceptron with multiple hidden layers process invariant features (distances, inverse distances, angles, Morse terms) with SIgmoid-Linear-Unit activations and skip connections and Radial Basis Function encodings. A 30\% energy-only warm-up is followed by combined optimization with gradual force weight ramping in the two-phase training. The force loss uses Huber loss with a threshold of 0.5. 
\end{itemize}

All methods undergo preprocessing and post-correction. Preprocessing is done identically across all models, including MinMax scaling for energies and forces, (fit only on training data to prevent information leakage). Post-correction entails a quadratic fitting for energies and a linear fitting for forces for addressing systematic biases in model predictions. To counteract training instabilities, QML methods use Gradient clipping with a maximum norm of 10.0 and the classical method a maximum norm of 5.

\begin{table}[h]
\centering
\caption{Architecture and parameter comparison of the four methods}
\label{tab:architecture}
\begin{tabular}{lccc}
\toprule
\textbf{Method} & \textbf{Qubits} & \textbf{Depth/Layers} & \textbf{Parameters} \\
\midrule
Rot. Equiv. QML & 6 & 6 & 80 \\
Non-Equiv. QML & 4 & 4 & 48 \\
Graph Perm. QML & 4 & 4 & 108 \\
Classical Equiv. NN & --- & 3 layers (128-128-64) & 27,073 \\
\bottomrule
\end{tabular}
\end{table}

\subsection{k-fold cross validation}
We compare the generalizability of the four methods for molecular physics learning with a k-fold Cross Validation (CV) framework. Specifically, k=5 is chosen i.e. the full dataset is split into 5 equally sized folds to judge performance statistically across the folds. Iterating over each fold, the model trains on the 4 other folds (80\% of full data) and is evaluated on the chosen fold (20\% of full data). Therefore, every data point serves as test data exactly over all iterations. The models are initialized with consistent random seeds across folds while varying them between folds to model real world variance. There are some method agnostic aspects of training (two-phase, preprocessing etc) as discussed in the previous subsection.

In each iteration over the chosen fold, multiple metrics $R^2$( Coefficient of Determination (R-squared)), MAE (Mean Absolute Error), RMSE (Root Mean Squared Error) are evaluated. Statistical summary accumulates results across all folds, computing means, standard deviations, coefficients of variation, and ranges to methodically characterize both performance and consistency. To evaluate generalizability in a scale-agnostic method, the coefficient of variation, defined as standard deviation divided by mean, is calculated. The definitions of the metrics for generalizability are as follows:

\begin{itemize}
    \item \textbf{Generalizability} (Coefficient of Variation): 
    \begin{equation}
        \text{CoV} = \frac{\sigma}{\mu}
    \end{equation}
    with $\sigma$ being the standard deviation and $\mu$ the mean of R$^2$ scores across folds. The Lower the CoV, the more consistent the performance is relative to the mean.
    
    \item \textbf{Consistency} (Standard Deviation):
    \begin{equation}
        \sigma = \sqrt{\frac{1}{k}\sum_{i=1}^{k}(x_i - \mu)^2}
    \end{equation}
    where $x_i$ is the R$^2$ score for fold $i$ and $k$ is the number of folds. A lower $\sigma$ indicating smaller deviations across folds is more desirable.
    
    \item \textbf{Stability} (Range):
    \begin{equation}
        \text{Range} = \max_i(x_i) - \min_i(x_i)
    \end{equation}
    The absolute difference between best and worst performance amng the folds. Predictions across different data splits are more stable when the range or stability is lower.
\end{itemize}

\section{Results}
We now present visualization plots for performance metrics ($R^2$, MAE) as well as generalizability metrics, across all folds.

\subsection{$LiH$ Molecule}
\begin{figure}[!ht]
\includegraphics[width=1.0\columnwidth]{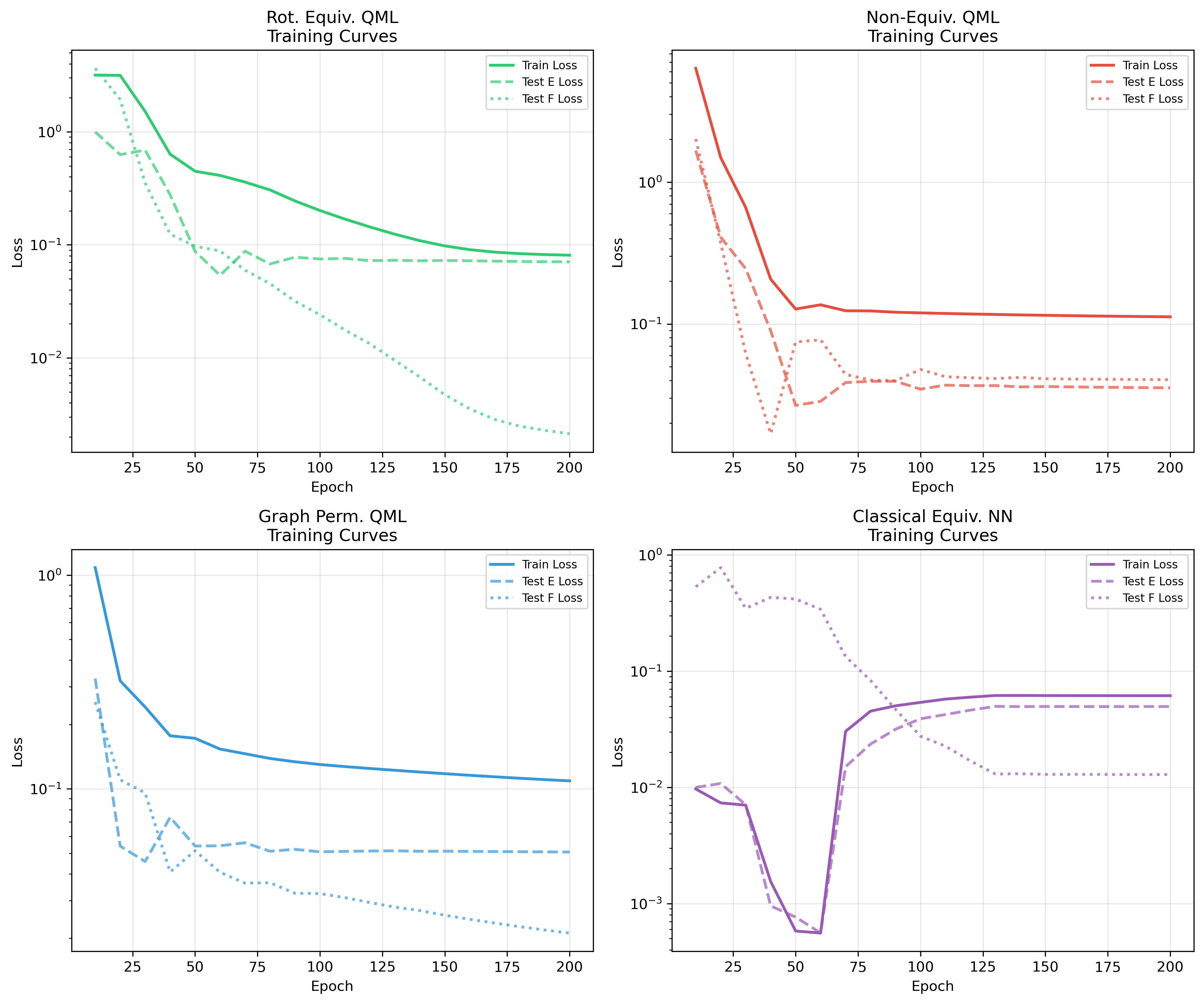}
\caption{\label{fig:LiH_training} Training of the four methods with $LiH$-data. All four methods use multi-phase training and/or adapting weights of energy and force in the cost function. }
\end{figure}

\begin{figure}[!ht]
\includegraphics[width=1.0\columnwidth]{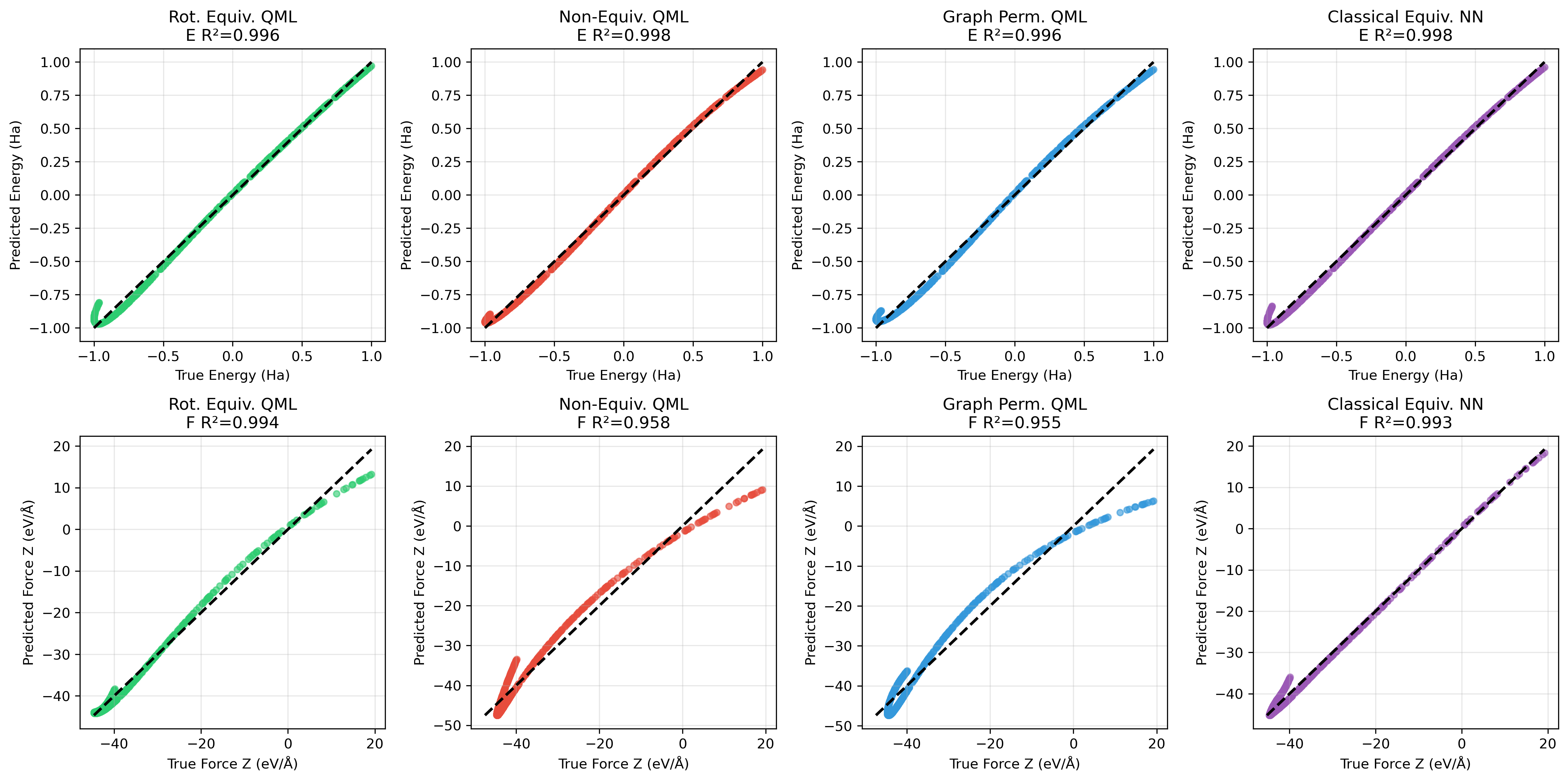}
\caption{\label{fig:LiH_preds} Energy and force predictions for the $LiH$-molecular data using the four methods.}
\end{figure}

With the multi-phase training schedule for energy and force, all four methods succeed in training over optimization of energy and force (Fig. \ref{fig:LiH_training}). Holistically (with the pre-processing and post-corrections), the classical method shows the best performance in the prediction task (Fig. \ref{fig:LiH_preds}). This is the overall picture. With the advancements of classical machine learning techniques, classical training is expected to succeed over any QML model. We are interested in the question of exactly how close we can come to the classical machine learning's performance with a moderate number of qubits, suitable of the capability of NISQ era. Since, we are using an equivariant classical network as well, we have set the bar quite high. The performance metrics from the classical network with the bias correction, is expected to be quite satisfactory. 

\begin{figure}[!ht]
\includegraphics[width=1.0\columnwidth]{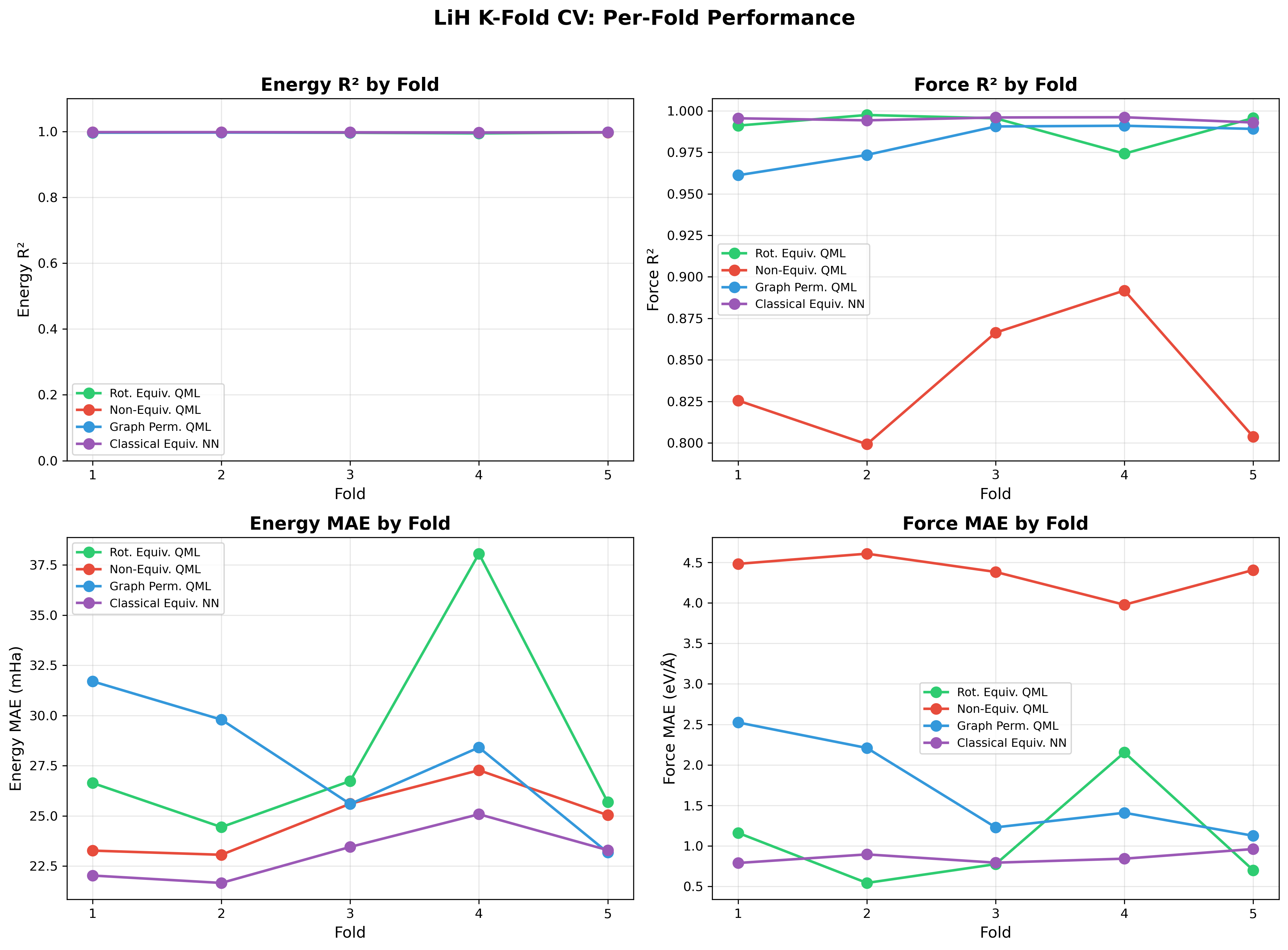}
\caption{\label{fig:LiH_kfold_perf} Performance of $LiH$-training Cross Validation: Energy and Force $R^2$ and MAE. }
\end{figure}

\begin{figure}[!ht]
\includegraphics[width=1.0\columnwidth]{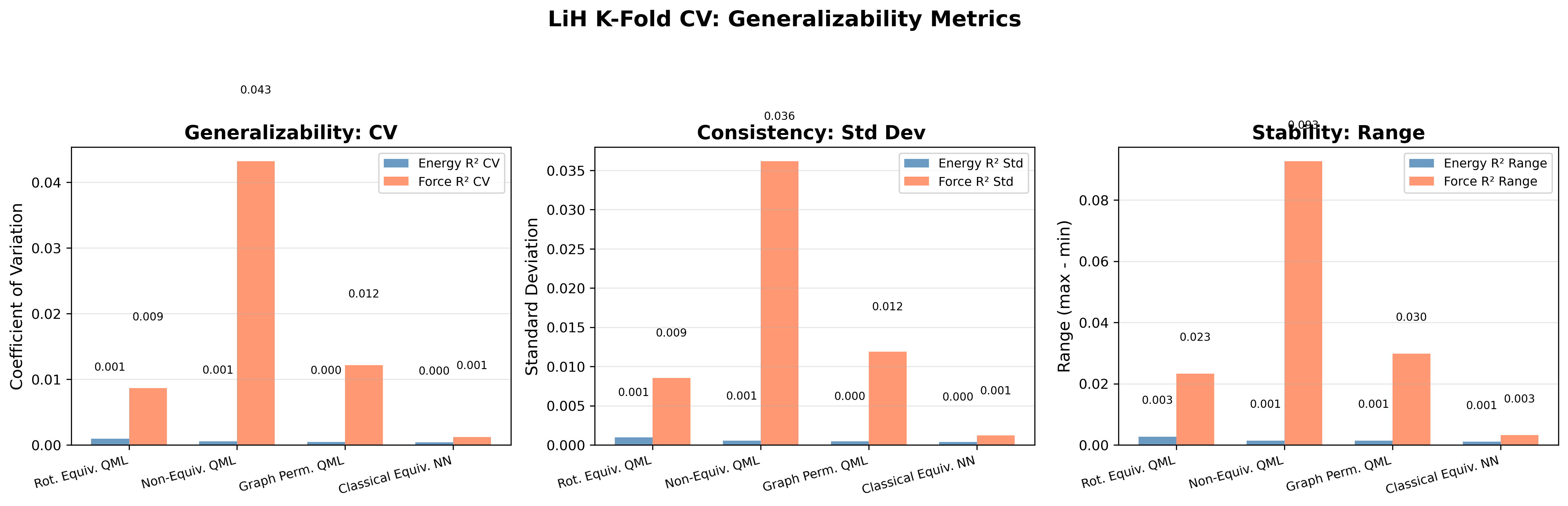}
\caption{\label{fig:LiH_genz0} Generalizability metrics of $LiH$-training Cross Validation: Energy and Force generalizability, consistency, and stability.}
\end{figure}

\begin{figure}[!ht]
\includegraphics[width=0.8\columnwidth]{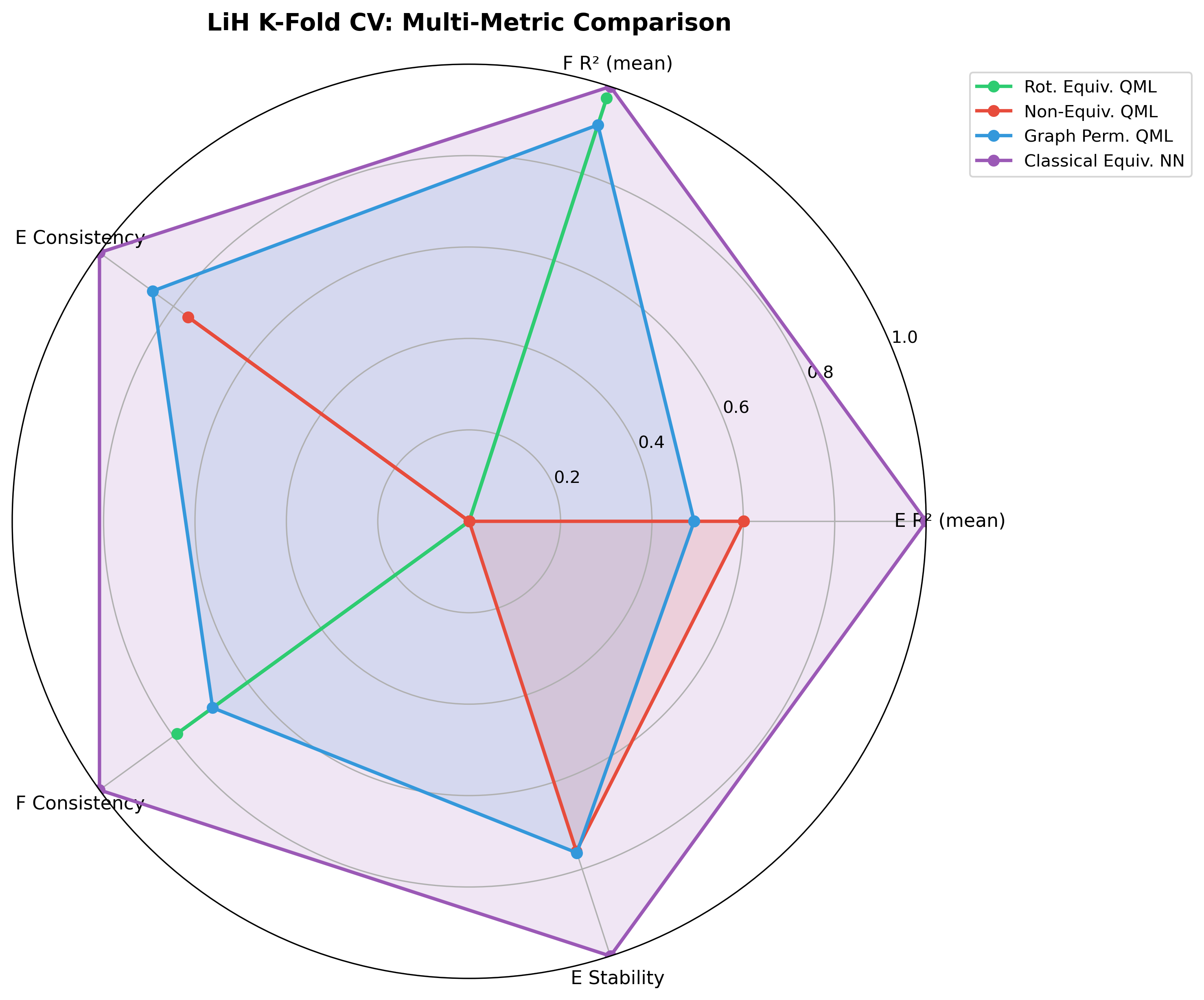}
\caption{\label{fig:LiH_radar} A Spider or Web chart for $LiH$: energy $R^2$, mean force $R^2$, energy consistency (inverted), force consistency(inverted), and energy stability(inverted).}
\end{figure}

In fact, the equivariance property is found to be quite critical. Fig.s  \ref{fig:LiH_kfold_perf}, \ref{fig:LiH_genz0}, \ref{fig:LiH_radar} show the trend that the non-equivariant network performs much worse than all the equivariant ones. The earlier statement of the classical machine learning method performing the best-- is borne out as well. The Spider plot in Fig. \ref{fig:LiH_radar} is the overall representation of this fact. The best performance comes from the purple pentagon where the classical method shows the maximum performance. The QML-methods compete against each other for the second best performance. Since, the non-equivariant QML model is the worst performer, the real competition is between the Rotationally Equivariant model and the Graph Permutaionally Equivariant model. 

Fig. \ref{fig:LiH_kfold_perf} shows that the competition between the two methods (Rotationally Equivariant and the Graph Permutaionally Equivariant model) can, in fact, be quite close. Over the folds, in terms of the metrics $R^2$ and MAE, one has no clear advantage or disadvantage over the other. Looking at Fig. \ref{fig:LiH_kfold_perf} only, one might conclude that the more computationally complex process of Graph permutational Equivariant model embedding may be unnecessary. If the more simplistic embedding of Rotationally Equivariant model can produce similar performance, it may be all we need. We remind here again that, some level of permutational invariance is in fact present in the Rotationally Equivariant method, in terms of the Heisenberg-type encoding (Eq. \eqref{eq_Two_Qub_Obs}). Is the embedding into a graph necessary, in that case? Fig. \ref{fig:LiH_genz0} gives us the answer to this question. While the Rotationally Equivariant model performs quite well, it somewhat lacks consistency or generalizability. The graph embedding in the method of Graph Permutationally Equivariant model is worth the effort if the added benefit of better generalizability is desired. The Graph Permutationally Equivariant model produces almost the same level of performance as Rotationally Equivariant model, and is consistent over its range. For an unknown or unseen data, we can trust the Graph Permutationally Equivariant model method more readily.

\subsection{$NH_3$ molecule}
For $NH_3$, we have three active H-atoms, and the graph embedding is more involved. A summary of the methodology is presented in Table \ref{table_NH3}.

\begin{table}[!ht]
\centering
\small
\begin{tabular}{|l|l|l|l|}
\hline
\textbf{Type} & \textbf{Feature (Our Model)} & \textbf{Molecular Input} & \textbf{Similar Ref.} \\
\hline

Node & Bond vectors $\mathbf{r}_{NH_i}$ & $\mathbf{x}_{H_i}-\mathbf{x}_N$ & \cite{batzner20223}, \cite{fang2022geometry} \\
Node & Distances $d_i=\|\mathbf{r}_{NH_i}\|$ & N--H bond lengths & \cite{batzner20223}, \cite{fang2022geometry} \\
Node & Unit vectors $\hat{\mathbf{r}}_{NH_i}$ & Normalized bond directions & \cite{batzner20223} \\
\hline

Edge & Angles $\theta_{ij}$ & $\angle(H_i-N-H_j)$ & \cite{fang2022geometry}, \cite{mao2025molecule} \\
Edge & Pairwise interactions & $(\mathbf{r}_{NH_i},\mathbf{r}_{NH_j})$ & \cite{batzner20223} \\
Edge & Cross-bond entanglement & Bond-pair connectivity & \cite{mao2025molecule} \\
\hline

\end{tabular}
\caption{ \label{table_NH3}Compact summary of geometric node and edge features in our NH$_3$ QML model and corresponding classical equivariant GNN references.}
\end{table}

\begin{figure*}[!ht]
\includegraphics[width=1.0\columnwidth]{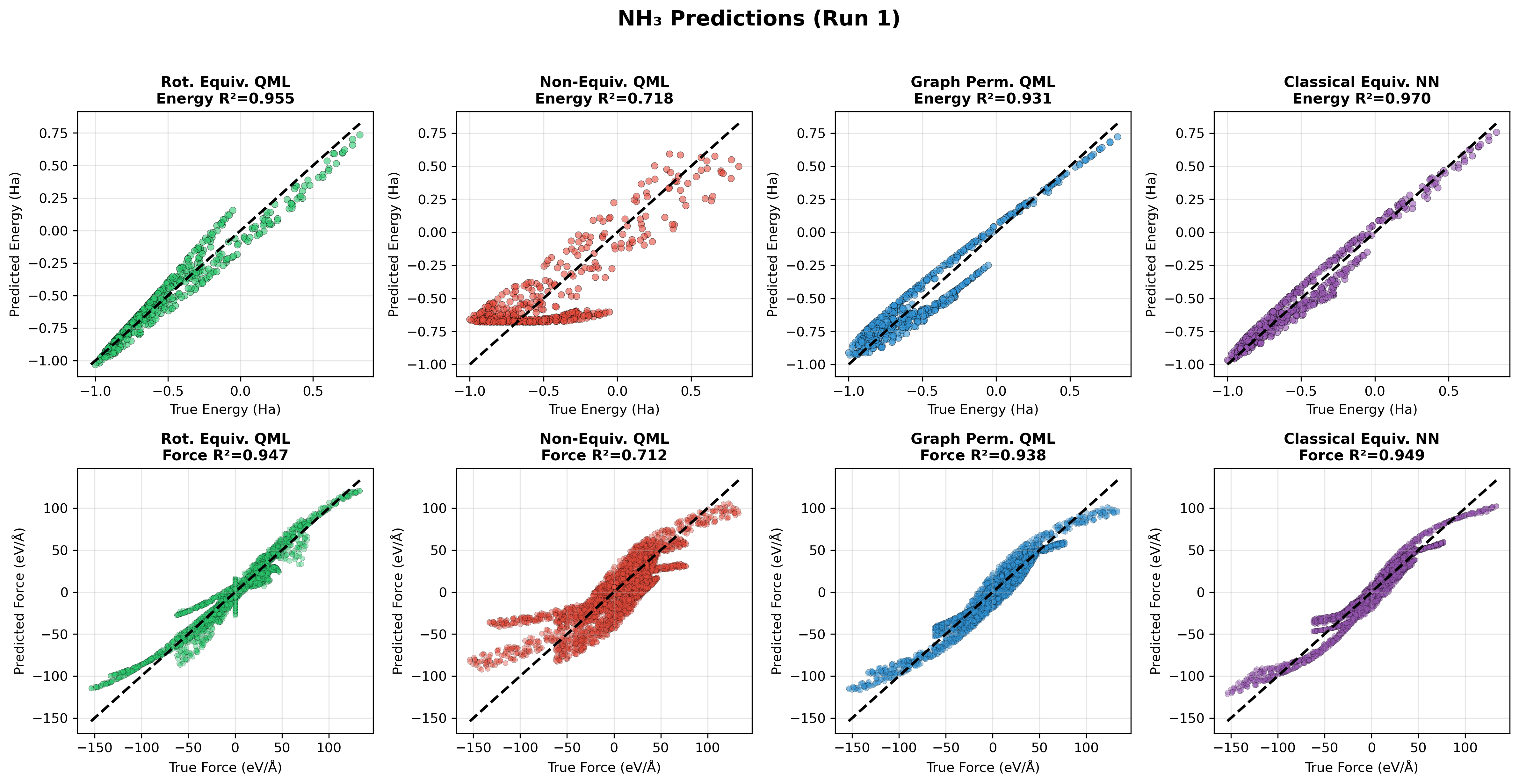}
\caption{\label{fig:NH3_preds} Energy and force (all components superimposed on one plot) predictions for the $NH_3$-molecular data using the four methods.}
\end{figure*}

\begin{figure}[!ht]
\includegraphics[width=1.0\columnwidth]{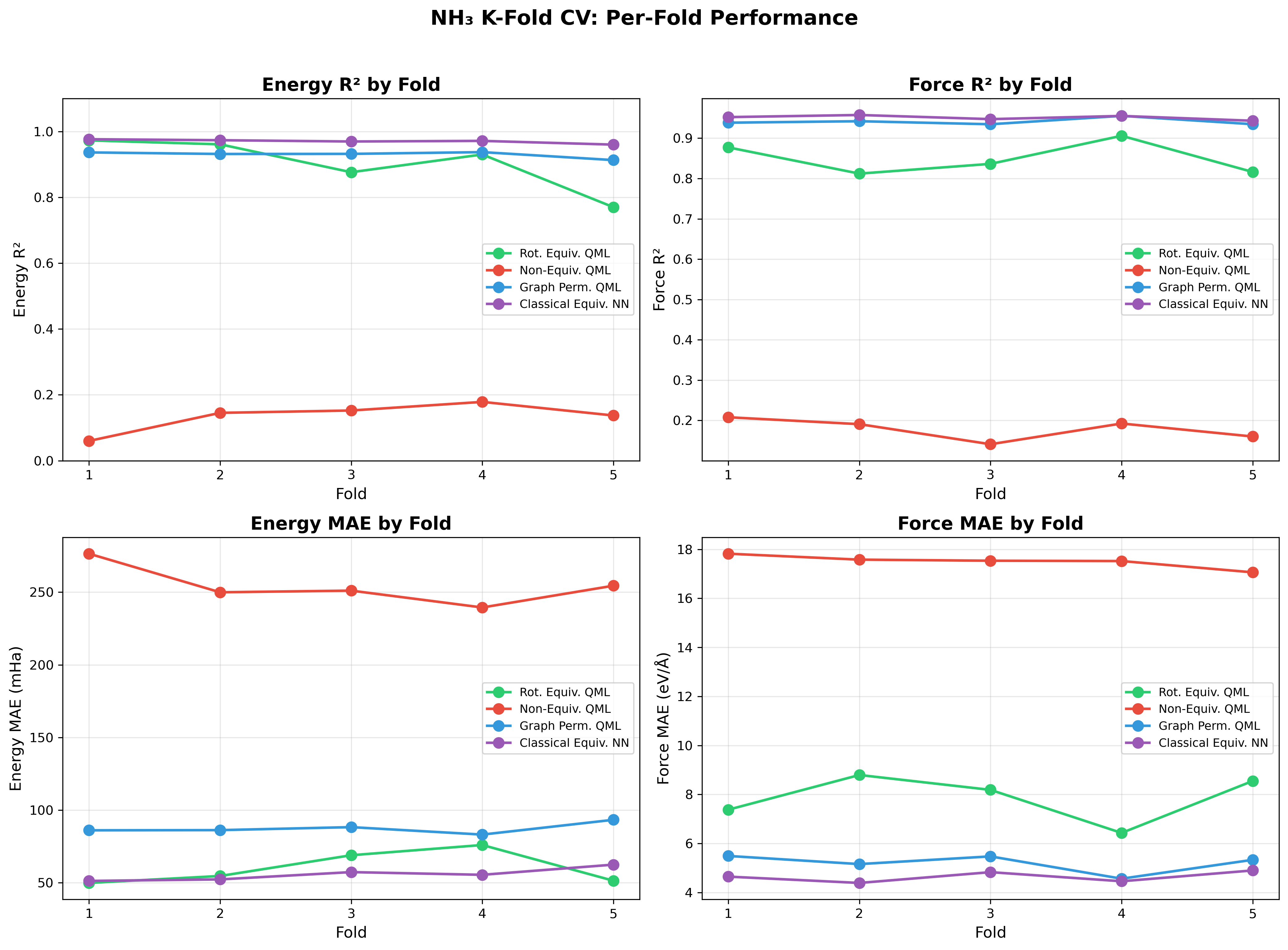}
\caption{\label{fig:NH3_kfold_perf} Performance of $NH_3$-training Cross Validation: Energy and Force $R^2$ and MAE. }
\end{figure}

\begin{figure}[!ht]
\includegraphics[width=1.0\columnwidth]{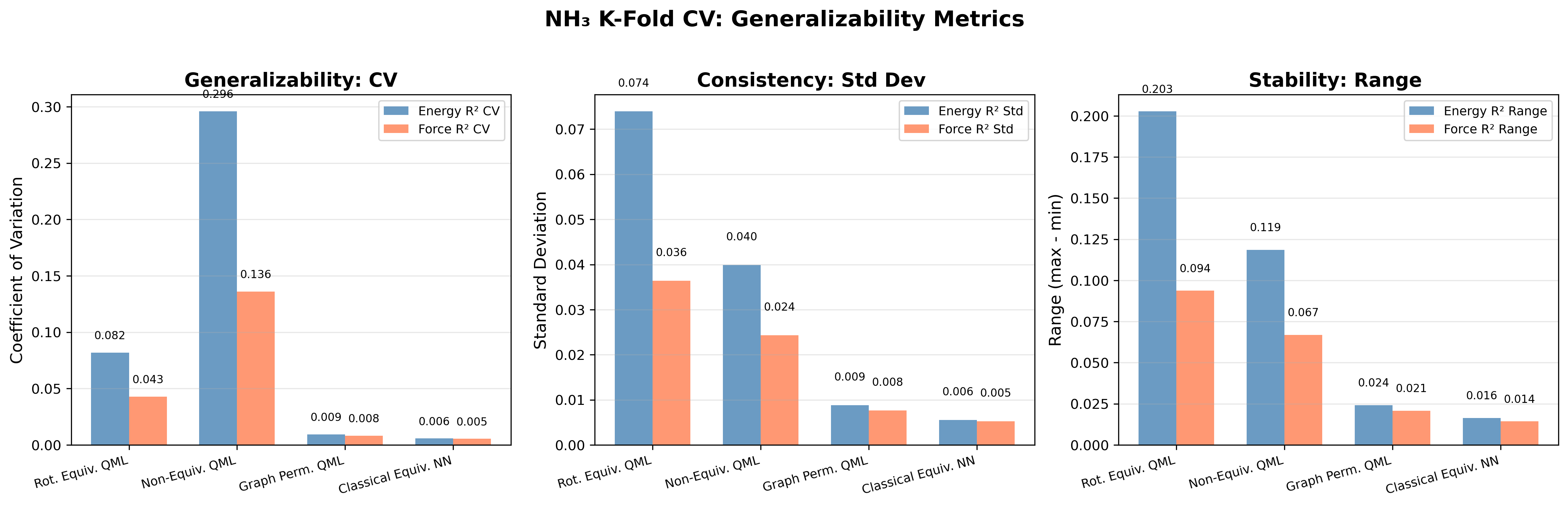}
\caption{\label{fig:NH3_genz0} Generalizability metrics of $NH_3$-training Cross Validation: Energy and Force generalizability, consistency, and stability.}
\end{figure}

\begin{figure}[!ht]
\includegraphics[width=0.8\columnwidth]{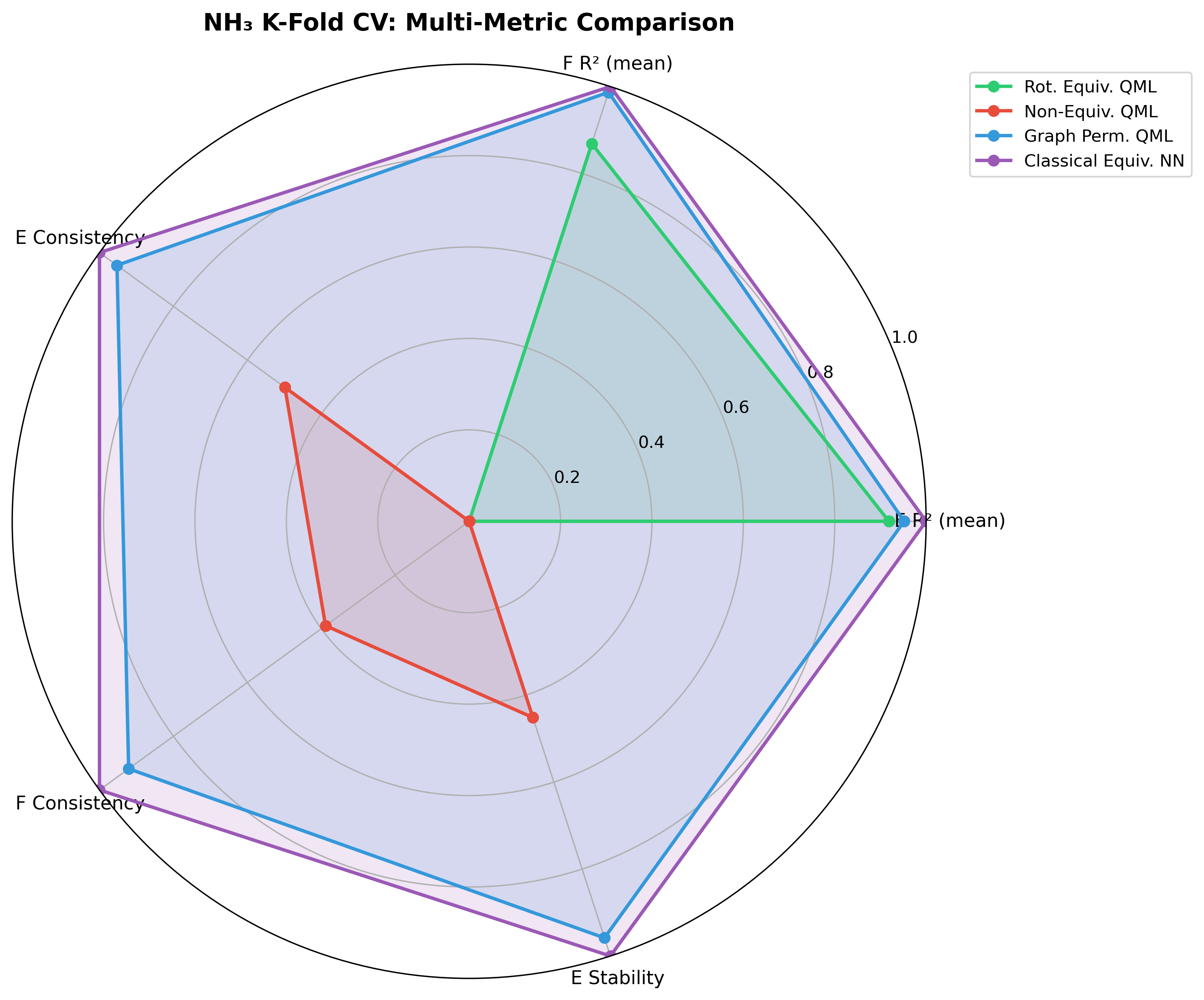}
\caption{\label{fig:NH3_radar} A Spider or Web chart for $NH_3$: energy $R^2$, mean force $R^2$, energy consistency (inverted), force consistency(inverted), and energy stability(inverted).}
\end{figure}

We do not show the training curves and results of the predictions (multidimensional for $NH_3$) componentwise for $NH_3$. But, they can be found in the online repository \cite{github_repo}. Fig. \ref{fig:NH3_preds} shows all prdicted force components superimposed on one plot and the energy predictions. Here again, the classical Equivariant model performs the best in terms of both accuracy and generalizability. The other conclusions from the $LiH$-data all hold true here as well. The worst method, once again is the QML with no equivariance. Now, there is an extra question about the competition between Rotationally Equivariant model and the Graph Permutaionally Equivariant model. Since the $NH_3$-molecule (trigonal pyramidal) has a much more interesting geometry than the $LiH$-molecule, is the generalizability is still comparable?

We now discuss the generalizability performance of the two methods (Rotationally Equivariant model and the Graph Permutaionally Equivariant model) for $NH_3$. While the performance metrics (Fig. \ref{fig:NH3_kfold_perf}) give no clear winner (the advantage could still be given to Graph Permutationally Equivariant model), the generalizability metrics (Fig.s \ref{fig:NH3_genz0}, \ref{fig:NH3_radar}) clearly has. As the problem now has more geometric aspects (compared to the linear $LiH$-molecule), the Graph Permutaionally Equivariant model generalizes much better, even almost catching up with the classical Equivariant model (Fig. \ref{fig:NH3_radar}). This is the important performance result of the paper. The generalizability of the training model is not independent of the geometry of the problem. If a molecular data has substantial geometrical imprint, a Graph Permutational Equivariant approach captures the physics of the problem much better. The advantage in generalizability metrics over a Rotaionally Equivariant model would be more obvious for a more geometry-heavy molecule. Among the QML models, a Graph Permutational Equivariant model should be preferred for a geometry-reliant physics problem.

\section{Concluding Remarks}
We have found clear advantages in GQML approaches over ordinary QML ones. When a problem with heavy geometrical attributes is involved, the choice of QML model should be cognizant of the symmetry properties of the data or at least the geometry. The generalization properties of QML depends critically on tapping into the equivariance properties. While equivariance can be chosen over some of the symmetries and not all, our results indicate that maximization of the symmetry equivariance leads to the most generalizable models. For a geometry-heavy problem, a graph embedding substantially improves the generalizability. A permutational equivariance assumption of the graph leads to much better trainability. The choice and distribution of geometrical and non-geometrical features into node and edge features merit further research.

On a numerical level, classical methods may always outperform QML methods. While it is expensive to increase the network size for QML (both at a simulator and hardware levels), the results presented in this paper show that there may be some benefit in using QML after-all, at least in terms of generalizability. When the dataset in question is geometric in nature, a Graph Permutationally Equivariant QML model gives us a way of tapping into the geometric attributes of the problem. Although, the classical machine learning parameters are inexpensive and a massive number of parameters (~27k in this work) can be thrown at it without much thought, our results do speak to the expressivity power of a QNN. If QNNs can produce almost comparable results with ~100 parameters (~300 fold reduction), they may have a prospect in learning complicated molecular processes with much fewer parameters. Immediate follow up work can attempt to challenge the QNN models with gradually increasing complexity.

The isomorphism of Lie Algebra $SO(3)$ and $SU(2)$ guarantees us that a QML problem can attempt a classical Equivariant model in the fashion of an exact map. As of now, the research goals should be emulating the performance of the classical machine learning only. But, in future, loftier aims may be possible. Molecular data is produced with quantum mechanical theories and quantum data is also a possibility. In the spirit of simulating quantum systems with a quantum computer, QML can take on the challenging task of learning from quantum data. The possibility of quantum advantage may not be a theoretical prospect forever. The tackling of a quantum dataset or similar revolutionary concepts may be tackled by the emerging field of QML.

\section*{Author Contribution Statement}
S.B. conceived of the research idea. S.B. and J.O. implemented the algorithms, performed the simulations, and conducted the comparative analysis. S.B. contributed to theoretical modeling, evaluation design, and result validation. Both authors wrote the manuscript, reviewed all sections, and approved the final version.

\section*{Funding}
This research did not receive any specific grant from funding agencies in the public, commercial, or not-for-profit sectors.

\section*{Data and Code Availability}
The results and datasets presented in this work can be reproduced with the code freely available from the repository \url{https://github.com/sbisw002/MoleQ-M-L.git}.

\begin{appendices}






\end{appendices}


\bibliography{EquiLiH}

\end{document}